\documentclass[10pt,twocolumn,letterpaper]{article}

\usepackage{iccv}
\usepackage{times}
\usepackage{epsfig}
\usepackage{graphicx}
\usepackage{amsmath}
\usepackage{amssymb}
\usepackage{float}
\usepackage{paralist}

\newcommand{\md}{\textsc{meta-dataset}}

\newcommand{\orbit}{\textsc{orbit}}

\newcommand{\mdshort}{\textsc{md}}

\newcommand{\attn}{\textsc{AttnScale}}
\newcommand{\attnlite}{\textsc{AttnScaleLite}}
\newcommand{\lnorm}{\textsc{LN-Tune}}

\newcommand{\linear}{\textsc{Linear}}
\newcommand{\protoaug}{\textsc{ProtoAug}}
\newcommand{\ncc}{\textsc{ProtoNCC}}

\usepackage{esrelation}
\usepackage{tikz}
\usepackage{graphics}
\usepackage{booktabs}
\usepackage{siunitx}
\usepackage{graphicx}
\usepackage{xcolor}

\newcommand*\circled[1]{\tikz[baseline=(char.base)]{
            \node[shape=circle,draw,inner sep=1pt] (char) {#1};}}
\usepackage[pagebackref=true,breaklinks=true,colorlinks,bookmarks=false]{hyperref}
\usepackage{amssymb}
\usepackage{cleveref}
\crefname{figure}{Fig}{Figs}%

\usepackage[ruled,vlined]{algorithm2e}
\definecolor{commentcolor}{RGB}{110,154,155}   % define comment color
\newcommand{\PyComment}[1]{\ttfamily\textcolor{commentcolor}{\# #1}}  % add a "#" before the input text "#1"
\newcommand{\PyCode}[1]{\ttfamily\textcolor{black}{#1}} % \ttfamily is the code font

\iccvfinalcopy % *** Uncomment this line for the final submission

\def\iccvPaperID{****} % *** Enter the ICCV Paper ID here

\ificcvfinal\fi
\begin{document}

%%%%%%%%% TITLE
\title{Strong Baselines for Parameter Efficient Few-Shot Fine-tuning}

\author{Samyadeep Basu$^{1}$, Daniela Massiceti$^{2}$, Shell Xu Hu$^{3}$, Soheil Feizi$^{1}$\\
$^{1}$University of Maryland, $^{2}$Microsoft Research, Cambridge,$^{3}$Samsung AI, Cambridge \\
{Correspondence to \tt\small sbasu12@umd.edu}
% \and
}

\maketitle
%\thispagestyle{empty}

%
%%% Author order: Samyadeep Basu, Shell Xu*, Daniela Massiceti*, Soheil Feizi with equal contribution from Shell and Daniela. 

%%%%%%%%% ABSTRACT
\begin{abstract}
\vspace{-0.10cm}
Few-shot classification (FSC) entails learning novel classes given only a few examples per class after a pre-training (or meta-training) phase on a set of base classes. Recent works have shown that simply fine-tuning a pre-trained Vision Transformer (ViT) on new test classes is a strong approach for FSC. Fine-tuning ViTs, however, is expensive in time, compute and storage. This has motivated the design of parameter efficient fine-tuning (PEFT) methods which fine-tune only a fraction of the Transformer's parameters. While these methods have shown promise, inconsistencies in experimental conditions make it difficult to disentangle their advantage from other experimental factors including the feature extractor architecture, pre-trained initialization and fine-tuning algorithm, amongst others. In our paper, we conduct a large-scale, experimentally consistent, empirical analysis to study PEFTs for few-shot image classification. Through a battery of over $1.8k$ controlled experiments on large-scale few-shot benchmarks including \md{} (\mdshort{}) and \orbit{}, we uncover novel insights on PEFTs that cast light on their efficacy in fine-tuning ViTs for few-shot classification. Through our controlled empirical study, we have two main findings: (i) Fine-tuning just the LayerNorm  parameters (which we call \lnorm{}) during few-shot adaptation is an extremely strong baseline across ViTs pre-trained with both self-supervised and supervised objectives, (ii) For self-supervised ViTs, we find that simply learning a set of scaling parameters for each attention matrix (which we call \attn{}) along with a domain-residual adapter (DRA) module leads to state-of-the-art performance (while being $\sim\!$ 9$\times$ more parameter-efficient) on \mdshort{}. Our extensive empirical findings set strong baselines and call for rethinking the current design of PEFT methods for FSC. 
%for few-shot learning in vision.
%Along the way we develop a simple and light-weight PEFT method called {\it AttnScale}, which learns a scaling factor for the attention weights in ViTs during few-shot adaptation (with $<1\%$ parameters). Through extensive experiments, we show that this extremely simple recipe reaches SoTA performances on existing challenging few-shot benchmarks. All of our code will be publically released.  
\end{abstract}
\vspace{-0.1cm}
%%%%%%%%% BODY TEXT
\section{Introduction}
%% general introduction about few-shot learning
% Few-shot classification is the ability to distinguish between a set of novel classes when given only a few labelled training examples of each class~\cite{one_shot, one_shot_new}. This holds potential across many real-world applications -- from robots that can identify new objects~\cite{open_world}, to drug discovery pipelines that can predict the properties of new molecules~\cite{stanley2021fsmol}. A few-shot {\it  image} classifier is given a few labelled training images of the new object classes, called the support set. Once the classifier has adapted to this support set, it is then evaluated on novel test images of those classes, called the query set. Together, the support and query set is called a task.
Few-shot classification (FSC) involves learning a new classification task given only a few labelled training examples from each of the novel classes. It has a large number of mainstream applications such as drug-discovery~\cite{stanley2021fsmol}, robotics~\cite{open_world} and personalized object recognition~\cite{orbit} among others. Usually, a given few-shot classification task consists of a few-labelled examples from the new classes (support set) and a testing set of unlabeled held-out examples of those classes (query set). 
 \begin{figure}
    \hskip -0.15cm
  \includegraphics[width=8.6cm, height=6.3cm]
  {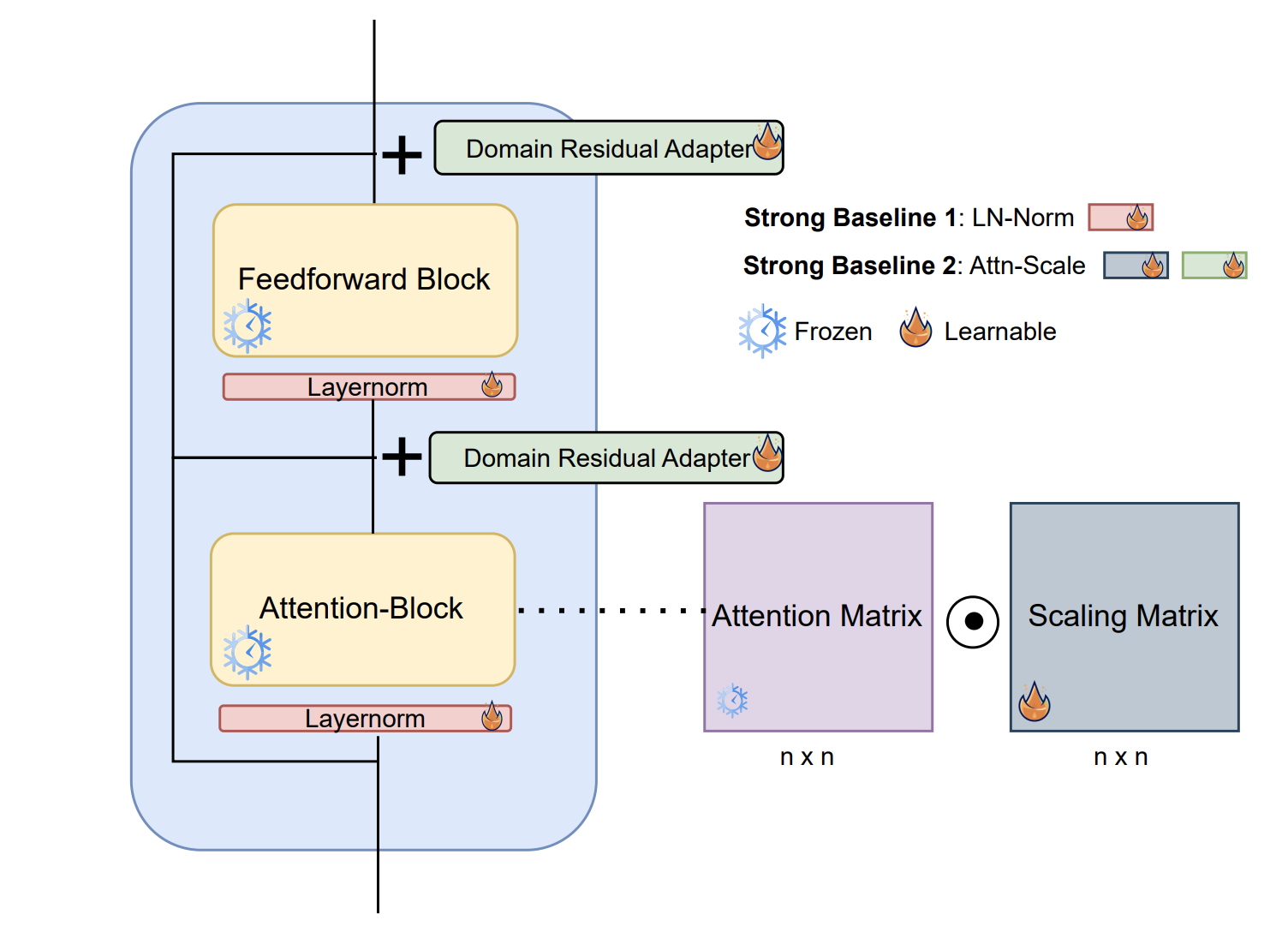}
 \caption{\label{main_diag} We introduce two {\it strong} PEFT baselines for few-shot image classification: (i) \lnorm{} which fine-tunes only the LayerNorm parameters of the ViT; (ii) \attn{} which fine-tunes a scaling parameter for the attention matrices along with a domain residual adapter. These approaches outperform full fine-tuning and all other existing PEFT methods on \mdshort{} and show competitive performance on \orbit{}. }
 \vspace{-0.3cm}
\end{figure}
\begin{figure*}
    \vspace{-0.8cm}
    \hskip 1.2cm
  \includegraphics[width=15cm, height=6.5cm]{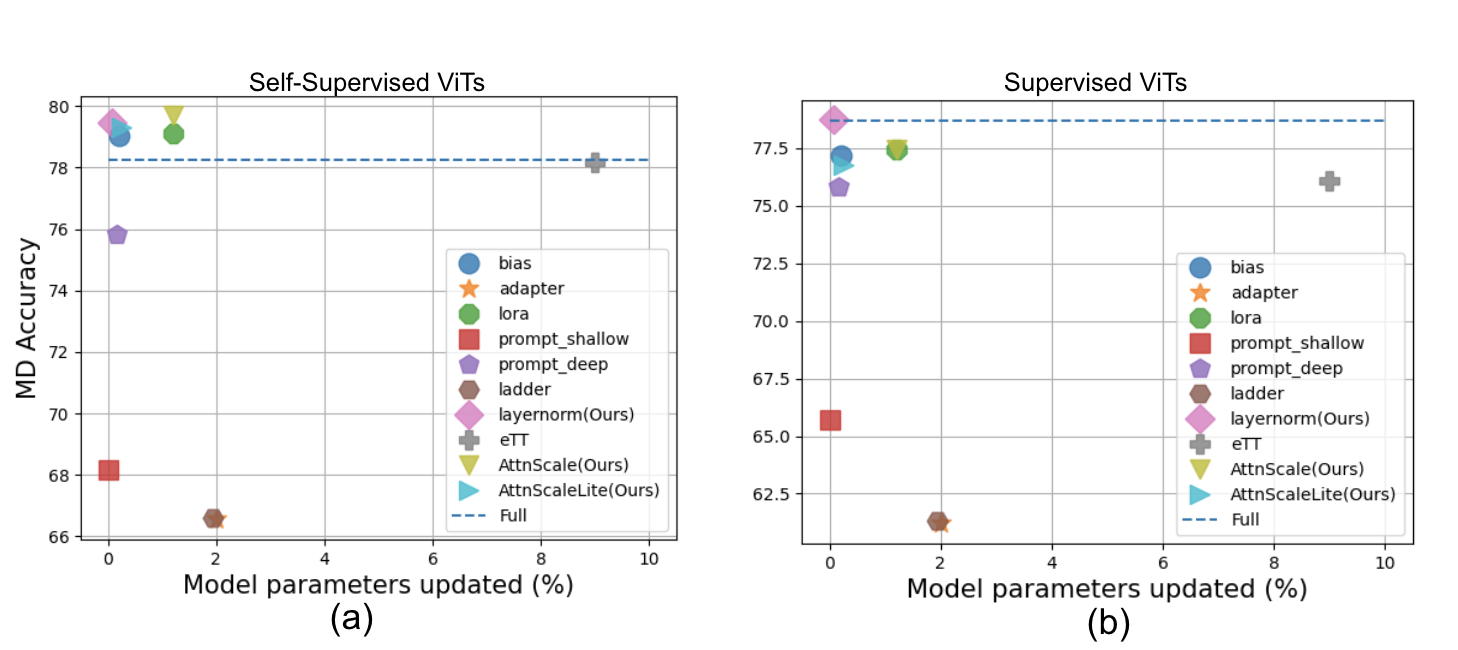}
  \vspace{-0.2cm}
    \caption{\textbf{\attn{} leads to SoTA performance on \mdshort{} with self-supervised ViTs and \lnorm{} leads to SoTA performance for supervised ViTs. } Pareto-Plot comparing the average \mdshort{} accuracy with the model parameters updated during few-shot adaptation:  (a) Averaged across self-supervised ViT-S/16 and ViT-B/16 (DINO); (b) Averaged across supervised ViT-S/16(DeiT), ViT-B/16(DeiT) and ViT-B/16(ImageNet-21k). We find that the recently proposed eTT~\cite{xu2022exploring} does not generalize well to supervised objectives and two simple but {\it strong} baselines \lnorm{} and \attn{} outperform existing PEFT methods. Averaged across all domains in \mdshort{} except ImageNet. }%
    \vspace{-0.50cm}
\end{figure*}

%% issue with full finetuning
Recent works~\cite{shell_simple_baselines, cross_task_adapters, xu2022exploring} have shown that fine-tuning a large pre-trained Vision Transformer (ViT) on the support set of new test tasks achieves state-of-the-art performance on large-scale few-shot classification benchmarks such as \md{} (\mdshort{}). Because of their high number of parameters, however, fine-tuning ViTs is extremely expensive in terms of storage, compute, and time. This limits the ability to learn new downstream tasks in real-world applications where resources are constrained (e.g., personalization on edge or mobile devices) since (i) storing the task's fine-tuned parameters on the edge may be unfeasible, especially for a large number of downstream tasks and (ii) fine-tuning on each new task takes prohibitively long. 

% Consider a few-shot classification system which is deployed to a large number of end users and needs to be personalized at the user's end. This scenario entails two major challenges: (i) First, fine-tuning all the parameters of the model will increase the storage cost significantly as the model copy needs to be stored for every user. (ii) Second, fine-tuning the entire model parameters comes with the disadvantage of an increase in personalization time. 
%% layland of peft methods for few-shot learning
As a result, much recent progress has been made in designing light-weight, fast and parameter-efficient fine-tuning (PEFT) methods~\cite{xu2022exploring, vtab_}. These reduce the computational requirements to adapt a ViT to a new test task by fine-tuning only a fraction of the ViT's total parameters. However, inconsistencies in experimental setups make it difficult to disentangle the benefit of PEFT methods from other experimental factors, including pre-training initialization, feature extractor architecture, fine-tuning algorithm, downstream dataset and other hyperparameters. Prompt-tuning~\cite{vtab_}, for example, is the state-of-the-art PEFT method on the transfer learning benchmark VTAB~\cite{vtab}, while eTT~\cite{xu2022exploring} performs strongly on few-shot classification in~\mdshort{}. Both, however, use distinct feature extractors, pre-training initializations, fine-tuning algorithms, and hyperparameters, thus limiting our understanding of the generalizability of these PEFT methods across different setups. 
% Moreover, the effectiveness of PEFT methods specifically for few-shot classification is still not well understood with~\cite{xu2022exploring} being the only work to test PEFT methods on the large-scale \md{} benchmark. 

% Existing PEFT solutions for few-shot learning in vision are either very specific to the pre-training paradigm, initialization or downstream dataset ~\cite{xu2022exploring, vtab_}. Moreover, different methods often use distinct hyper-parameter tuning and model selection strategies, thereby rendering a fair and realistic comparison of these PEFT methods difficult. 

% what we do in our paper - part (i) -- empirical study
To address this, we perform a large-scale empirical analysis of top-performing PEFT methods on two large-scale few-shot image classification benchmarks, \md{}~\cite{metadataset} and \orbit{}~\cite{orbit}. Our experimentation involves $\sim\!\!1.8k$ fine-tuning experiments which quantify the performance of PEFT methods under experimentally controlled settings including ViT architectures, pre-training objectives, and fine-tuning algorithms. This enables us to compare PEFT methods in a fair and consistent way and also draw out novel insights on the interaction between these different components in the fine-tuning pipeline.

% The scale of our experiments unravel various insights about the limitations of existing PEFT methods. For e.g., (i) In our controlled experimental test-bed, we show that none of the carefully designed existing PEFT methods show consistent performance across different pre-training objectives. (ii) We find that for different domain shifts, distinct PEFT methods are preferred highlighting that there is no one 'winner' PEFT module. (iii) Dropping PEFT methods from certain layers without sacrificing accuracy strongly depend on the downstream domain to which the model is adapting. 
Our main finding is that the embarrassingly simple approach of fine-tuning just a ViT's LayerNorm parameters (only 0.08$\%$ of total parameters) on a new test task leads to better performance than with full model fine-tuning and other PEFT methods on \mdshort{} and \orbit{}. We call this baseline \lnorm{}. We also find that the recently proposed eTT~\cite{xu2022exploring}, primarily designed for self-supervised ViTs, lags behind some of the PEFT methods which we evaluate in our empirical study. In lieu of this, we propose a new strong baseline called \attn{} which leads to improved few-shot performance over eTT and other PEFT methods for self-supervised ViTs. In particular, \attn{} learns only a scaling parameter for each entry in the attention matrices along with a domain-residual module during few-shot adaptation, making it $\sim\!$ 9x more parameter-efficient than eTT. Importantly, \attn{} is extremely simple to implement, requires less than 6 lines of code, and can be easily integrated with {\it any} ViT architecture. 
%With this -- we establish two strong PEFT baselines for few-shot classification.  

These approaches establish two new, strong PEFT baselines for few-shot classification, however our empirical study also reveals several interesting insights: (i) None of the carefully designed existing PEFT methods show consistent performance rankings across different pre-training methods (\Cref{ref_pretrain}). (ii) We find that for different degrees of domain shifts, distinct PEFT methods are preferred highlighting that the need for surgically designing PEFT methods for different domain shifts (\Cref{consistencydomains}). (iii) Dropping PEFT methods from earlier layers in the ViT for large domain shifts (e.g. Omniglot, Quickdraw, Traffic-Sign) is detrimental to few-shot performance (\Cref{ref_drop}).
In summary, our contributions are as follows: 

\begin{compactitem}
    \item A large-scale, experimentally consistent, empirical analysis of a wide-range of PEFT methods for few-shot classification on 2 challenging large-scale benchmarks, \md{} and \orbit{}. 
    % \item A new embarrassingly simple baseline -- \lnorm{} which shows better few-shot performances than existing PEFT methods using only 0.08$\%$ 
    % \item Two embarrassingly {\it simple} but {\it strong} parameter-efficient few-shot fine-tuning methods called \lnorm{} and \attn{} which can fine-tune ViTs with less than $0.08\%$ and $1.2\%$ of the entire network parameters respectively with performance better than full model tuning and existing PEFT methods. 
    \item An embarrassingly simple PEFT baseline, \lnorm{}, which fine-tunes less than 0.08$\%$ of a ViT's parameters outperforming all existing PEFT methods on \mdshort{} amongst supervised ViTs.
    \item An easy-to-implement method, \attn{}, which sets a new state-of-the-art on \mdshort{} amongst self-supervised ViTs while fine-tuning $<\!1.2\%$ of the ViT's parameters. 
\end{compactitem}
% major takeaways and research progress from our paper
Our findings highlight that there is no one-size-fits-all PEFT method and simple parameter-efficient fine-tuning baselines should not be overlooked.  
%For e.g., \cite{xu2022exploring} design a attentive-prefix tuning module for ViTs initialized with DINO~\cite{dino}, while \cite{shell_simple_baselines} design a data-augmentation based fine-tuning technique for a variety of pre-trained ViTs. 

\section{Related Works}
\textbf{ViTs in few-shot classification.} CNNs have primarily been used as the feature extractor backbone in few-shot classification methods~\cite{maml, protonets, meta_baseline, meta_learning_survey, matching_networks}, however, recently ViTs have replaced them as the state-of-the-art~\cite{shell_simple_baselines} in challenging few-shot classification benchmarks like~\md{}. In these methods, the ViT is typically pre-trained with a self-supervised (or meta-learning) objective on a large dataset and then fine-tuned on new test tasks. While some works~\cite{sun, hiller2022rethinking} have explored pre-training techniques to make ViTs specifically suited to downstream few-shot classification, fine-tuning a ViT at test time remains expensive. Our work therefore aims to shed light on parameter-efficient fine-tuning methods for few-shot classification. 

\textbf{PEFT methods for few-shot classification.} Parameter efficient fine-tuning methods have been extensively studied in Transformers for NLP tasks with adapters~\cite{adapter}, LoRA~\cite{lora}, prefix-tuning~\cite{prefix_tuning} and prompt-tuning~\cite{prompt_tuning_nlp} serving as strong alternatives to fine-tuning all the Transformer's parameters. PEFTs have also been explored in Vision Transformers for computer vision tasks, with methods like visual prompt tuning~\cite{vtab_} for transfer learning which work by tuning prefixes attached to the input and eTT~\cite{xu2022exploring} which tune prefixes attached to key and value matrices in the self-attention layers. \cite{xu2022exploring} show that eTT results in performance close to full model tuning for ViTs pre-trained using DINO using only 9$\%$ of the total model parameters on the large-scale \md{}.  

\vspace{-0.2cm}
\section{Few-Shot Classification Preliminaries}
%% overview of the problem  + definition of the problem in equations
In few-shot classification, the goal is to adapt a classifier to a new task at test time using a small number of training examples of each new class. In fine-tuning-based approaches, this adaptation process is done by fine-tuning the model on the training examples, before then evaluating it on a held-out set of test examples.

Formally, given a pre-trained feature extractor $f_{\theta}$, a few-shot task is sampled from a test dataset $\mathcal{D}$. The task is composed of a support set $\mathcal{S}$ (of training examples) and a query set $Q$ (of held-out test examples). Generally, $N$ unique classes are first sampled from the underlying dataset $\mathcal{D}$. For each class $j \in [1, N]$, $k_{s}^{j}$ examples are sampled for the support set $\mathcal{S}$ and $k_{q}^{j}$ examples are sampled for the query set $\mathcal{Q}$.  If $k_{s}^{j}=k$ is fixed for $\forall j \in [1,N]$ classes, then the task is known as a $N$-way, $k$-shot task. When given a new test task, the objective is to fine-tune the underlying feature extractor $f_{\theta}$ or the parameter-efficient module $p_{\phi}$ on the task's support set $\mathcal{S}$ using a fine-tuning algorithm $\mathcal{F}$.  In parameter-efficient fine-tuning approaches, $f_{\theta}$ is frozen and only the parameters in $p_{\phi}$ are fine-tuned. More specifically, we can formalize the fine-tuning procedure as follows:
\begin{align}
    \label{maineq}
    \phi^{*} = \min_{\phi } \ell(f_{\theta}, p_{\phi}, \mathcal{F}(\mathcal{S}))
\end{align}
Inference on the query examples is done depending on the fine-tuning algorithm $\mathcal{F}$ (see~\Cref{ft_strategy}) for details). We follow the variable-way, variable way sampling protocol from~\cite{metadataset} where $k_{s}^{j}$, $k_{q}^{j}$ and $N$ vary for each sampled few-shot task. This setting generates class-imbalanced few-shot tasks which make it challenging as the model needs to handle tasks of varying sizes. 
\begin{figure*}
    \vspace{-0.8cm}
    \hskip 1.4cm
  \includegraphics[width=14.5cm, height=7cm]{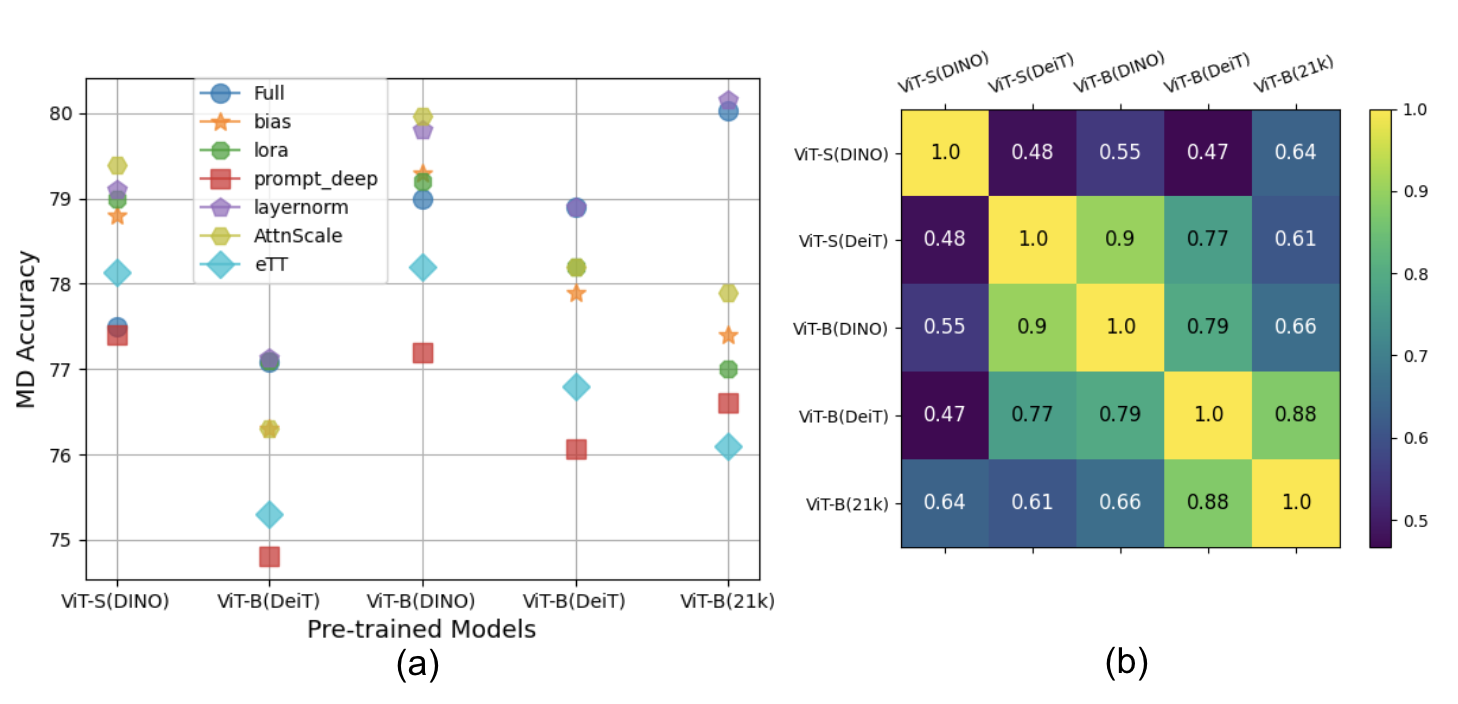}
  \vspace{-0.3cm}
    \caption{\label{pretrain_consistency} \textbf{PEFT methods (except our \lnorm{} strong baseline) lack consistency across different pre-training paradigms.} (a) The ranks of the 7 top-performing PEFT methods on \md{} change across different pre-training paradigms when measured under controlled settings; (b) The Spearman correlations between the different pre-trained models with respect to the performance rank of all 10 PEFT methods are not consistently high. Evaluation across all domains in \mdshort{} except ImageNet.}%
    \vspace{-0.50cm}
\end{figure*}
\section{Large-Scale Empirical Study Design}
\label{exp_test_bed}

PEFT methods have been widely used to make few-shot adaptation more computationally efficient~\cite{vtab_, xu2022exploring, fit}, however, inconsistencies in experimental setups make it difficult to disentangle the gain from PEFT methods versus other experimental factors. To address this, we conduct a wide-scale experimentally controlled study of over $1.8k$ experiments. We control for the pre-trained model (including pre-training objective and architecture), PEFT module type, position of the PEFT module, fine-tuning algorithm, learning hyperparameters and downstream dataset. Below we provide details of each of these components:

\textbf{Pre-trained models}. For pre-training objectives we consider the self-supervised objective DINO~\cite{dino} and the supervised objective DeiT~\cite{deit}. For architectures, we consider ViT-S/16 and ViT-B/16~\cite{deit}. These architectures are pre-trained using the given objectives on ImageNet-1k. In addition, we also consider ViT-B/16, which is pre-trained on the large-scale ImageNet-21k. These objectives and architectures were chosen as they lead in downstream few-shot performance~\cite{shell_simple_baselines} on \mdshort{}. More details on pre-training are included in the Appendix. 
% We consider ViT-S/16 (DINO), ViT-B/16 (DINO), ViT-S/16 (DeiT), ViT-B/16 (DeiT), and ViT-B/16 (ImageNet-21k). For self-supervised objectives we choose DINO~\cite{dino} with ViT-S/16 and ViT-B/16. For supervised objectives, we choose DeiT~\cite{deit} with ViT-S/16 and ViT-B/16 which are pre-trained on ImageNet-1k. We also test the effectiveness of large-scale pre-training on PEFT methods with ViT-B/16 pre-trained on ImageNet-21k with the DeiT supervised pre-training objective. We select these models as they are strongly performing vision backbones for few-shot classification~\cite{shell_simple_baselines}.

% ViT-S/16 and ViT-B/16 models with DINO~\cite{dino} and DeiT~\cite{deit} are pre-trained on ImageNet-1k. ViT-B/16 (ImageNet-21k) is pre-trained using the DeiT~\cite{deit} supervised pre-training strategy on ImageNet-21k.

\textbf{PEFT methods}.  \label{peft_description} We consider the following 7 existing methods for parameter-efficient fine-tuning: adapters~\cite{adapter}, LoRA~\cite{lora}, shallow prompt-tuning and deep prompt-tuning~\cite{vtab_}, eTT~\cite{xu2022exploring}, ladder tuning~\cite{ladder}, and bias tuning~\cite{bitfit}. We also compare to full model fine-tuning~\cite{shell_simple_baselines} and our 2 strong baselines: fine-tuning only the ViT's LayerNorm parameters (\lnorm{}), and learning a simple scaling factor for the elements in the attention matrices (\attn{}) (see~\Cref{attn_scale_description}). Of the existing methods, adapters and LoRA have been extensively used for fine-tuning Transformers in few-shot NLP tasks. Ladder tuning is a more recent memory-efficient as well as parameter-efficient fine-tuning method for language models like T5~\cite{t5}. Ladder is tuning is memory-efficient as it avoids back-propagation through the entire feature-extractor backbone. Shallow and deep prompt tuning are adaptations of~\cite{prompt_tuning_nlp} for transfer learning in vision. eTT~\cite{xu2022exploring} fine-tunes only the prefixes attached to the key and value matrices in a ViT's self-attention layers. eTT is also the only method to have been tested on the large-scale \md{} benchmark. Note, we omit the prototype regularization used in eTT to ensure fair comparison to other PEFT methods where prototype regularization is not used. We provide further information for each of these methods in the Appendix.

% \textbf{Position of PEFT methods.} We consider ViTs with the PEFT inserted in each of the layers, including the final. With 12 layers in both ViT-S/16 and ViT-B/16, each fine-tuning experiment is also repeated 12 times (see~\Cref{ref_drop} for analyses).
\textbf{Position of PEFT methods.} We consider two configurations in which the PEFTs are inserted in the ViT: (i) We insert PEFTs in each of the layers, including the final; (ii) We insert PEFT in the final layer and in one of the layers between the first and the final layer, leading to two layers in total. For (ii) each fine-tuning experiment is repeated 12 times (see~\Cref{ref_drop} for analyses).

\textbf{Fine-tuning algorithms \label{ft_strategy}}. We consider 3 fine-tuning algorithms given a new test task: (i) \linear{}: We attach a linear classification layer after the final layer of the ViT and fine-tune both the PEFT's and this layer's parameters using a cross-entropy loss. (ii) \protoaug{}: Following the state-of-the-art fine-tuning approach in~\cite{shell_simple_baselines}, we use the examples from the task's support set to initialize class prototypes, similar to ProtoNets~\cite{protonets}, and then use a query set to fine-tune the ViT. where the query set is an augmented version of the support set. In particular, we apply color-jitter and translation augmentations on the support set to generate the query set. (iii) \ncc{}: Following ~\cite{cross_task_adapters, xu2022exploring}, we do not apply augmentations to generate the query set and instead treat the query set as a copy of the support set, and fine-tune the ViT in a similar way to \protoaug. For (ii) and (iii), inference on the query set is performed using a ProtoNets classifier~\cite{protonets}, while for (i), the linear classifier is used. 

\textbf{Hyperparameters}. \label{model_sel} We standardize the hyperparameters across our entire experimental setup. Following~\cite{shell_simple_baselines}, we choose a learning rate from $\{0.0001, 0.001, 0.01, 0.1\}$ and select the rate that gives the best performance on the validation set. The validation set is a fixed set of 5 few-shot tasks sampled from the downstream dataset to which the ViT is being adapted. For each few-shot task, we fine-tune for 40 steps with the Adam optimizer~\cite{adam} using the selected learning rate. 
% In particular, we follow~\cite{shell_simple_baselines} to select the learning rate. For each sub-dataset in \md{}, we pre-select 5 few-shot tasks which we treat as the validation set. We choose a learning rate from $\{0.0001, 0.001, 0.01, 0.1\}$ which results in the best performance on the validation set. 

\textbf{Downstream datasets}.  We run all our experiments on two challenging large-scale few-shot classification benchmarks (i) \md{}~\cite{metadataset} and (ii) \orbit{}~\cite{orbit}. \md{} consists of 10 different sub-datasets, and is currently the most widely used few-shot classification benchmark. Note, we remove the ilsvrc$\_$2012 sub-dataset from \md{} as our ViT models have been pre-trained on ImageNet. \orbit{} is a few-shot classification benchmark containing noisy, real-world videos of everyday objects across 17 test users. In accordance with~\cite{metadataset}, we sample 600 few-shot tasks per sub-dataset in \md{} while for \orbit{}, we sample 50 tasks per user. In total, each experimental analysis is performed on 6250 few-shot tasks. 

\textbf{GPU compute.} Given the large memory requirements to fine-tune ViTs especially for tasks sampled from \mdshort{} (due to large support set sizes), we use an A6000 GPU (with 48GB memory) for ViT-B/16 and an A5000 GPU (with 24GB memory) for ViT-S/16.  
% \textbf{Fine-tuning hyperparameters. } For each few-shot task, we fine-tune for 40 steps with the Adam optimizer~\cite{adam} using the learning rate selected by the model selection criterion in \Cref{model_sel}. For \protoaug{} fine-tuning algorithm, we apply color-jitter and translation augmentations to the support set to generate the query set.  
\vspace{-0.2cm}
\section{Embarrassingly Simple Strong Baselines for Few-Shot Fine-tuning}
Our standardised large-scale empirical study led us to discover two embarrassingly simple but strong baselines for parameter-efficient few-shot fine-tuning: \lnorm{} and \attn{}. Both of these methods perform better than full model fine-tuning and all other existing PEFT methods on \mdshort{} at a fraction of the computational cost. Below we describe each of these strong baselines: 
% \lnorm{} works by fine-tuning {\it only} the ViT's LayerNorm parameters on a task's support set. We find that is an extremely strong baseline for ViTs pre-trained with both self-supervised and supervised objectives, beating full fine-tuning and all existing PEFT methods and tuning less than 0.8\% of the ViT's parameters.  
% In Sec. (\ref{empirical}), we find that for self-supervised pre-training objectives such as DINO - PEFT methods which directly impact the attention block in the ViT (e.g., eTT~\cite{xu2022exploring}), result in strong few-shot performances. With this observation, we propose an embarrassingly simple baseline for self-supervised ViTs called \attn{} which directly learns a scaling factor for the attention matrix during few-shot adaptation using the support set. In particular \attn{} replaces the attentive prefix tuning part off eTT~\cite{xu2022exploring} with a learnable scaling parameter on the attention matrix, which is tuned along with the domain residual adapter module from eTT. This modification results in 9x lesser parameters for few-shot fine-tuning than eTT, with better performances on \md{}.  
\vspace{-0.05cm}
\subsection{\lnorm{}}
\lnorm{} works by fine-tuning {\it only} the ViT's LayerNorm parameters on a task's support set. Formally, for a given ViT with $L$ layers, the $i^{th}$ layer has two LayerNorm blocks -- one before its attention block and one before its MLP block. Given an input vector $a \in \mathbb{R}^{d}$ from the previous layer or block, the operation of the first block can defined as LayerNorm$^{i}_{1}$(a) $= \gamma_{1}^{i} \odot (a-\mu)/\sigma + \beta_{1}^{i}$, and the operation of the second block as LayerNorm$^{i}_{2}$(a) $= \gamma_{2}^{i} \odot (a-\mu)/\sigma + \beta_{2}^{i}$. Here $\{\gamma_{1}^{i}, \beta_{1}^{i}, \gamma_{2}^{i}, \beta_{2}^{i}\} \in \mathbb{R}^{d}$ are the only learnable parameters for the $i^{th}$ layer. For a given task, these parameters across all $L$ layers are fine-tuned using the task's support set $\mathcal{S}$. As a result, \lnorm{} is extremely light-weight when compared to the other PEFT methods. For e.g., a ViT-S/16 has only $\sim\!18.6k$ LayerNorm parameters, while a ViT-B/16 has only $\sim\!37k$. Since ViT-S/16 and ViT-B/16 have $\sim\! 22$M and $\sim\! 76$M parameters, respectively, this accounts for less than $0.08\%$ of the total parameters. 
\vspace{-0.1cm}
\subsection{\attn{}}
\label{attn_scale_description}
As a second strong baseline, we introduce \attn{}, a modification to the recently proposed eTT~\cite{xu2022exploring}. Here, we replace the attentive prefix tuning part in eTT with a learnable scaling parameter on each element in the attention matrices, which we tune along with eTT's DRA module, reducing the number of learnable parameters by $\sim\!$ 9x.  
% While eTT was primarily designed for self-supervised ViTs, we find that \attn{} results in strong performances also on supervised ViTs pre-trained on both ImageNet-1k and ImageNet-21k. 
Given a ViT with $L$ layers, $n_{h}$ attention heads and $n$ tokens, 
% we define a learnable scaling tensor $A_{\alpha} \in \mathbb{R}^{n \times n \times L \times n_{h}}$. $A_{\alpha}$ can be reshaped as $\{ A_{\alpha}^{i} \}_{i=1}^{L}$ where $A_{\alpha}^{i} \in \mathbb{R}^{n \times n \times n_{h}}$ is the scaling tensor for each $i^{th}$ layer. For each attention head $j\in [1,n_{h}]$, the scaling matrix is defined as $A_{\alpha}^{ij} \in \mathbb{R}^{n \times n}$. 
the weight matrices in the $i^{th}$ layer's attention block for the $j^{th}$ head are defined as $W_{q}^{ij} \in \mathbb{R}^{d \times d_{e}}$, $W_{k}^{ij} \in \mathbb{R}^{d \times d_{e}}$ and $W_{v}^{ij} \in \mathbb{R}^{d \times d_{e}}$. Here $d$ is the dimension of the token embeddings and $d_{e}$ is the dimension of the tokens after the weight matrix projection. $Q^{ij} \in \mathbb{R}^{n \times d}, K^{ij} \in \mathbb{R}^{n \times d}, V^{ij} \in \mathbb{R}^{n \times d} $ are defined as the query, key and value tokens, respectively. The attention matrix in the $i^{th}$ layer for the $j^{th}$ head can be defined as: 
\begin{align}
    A^{ij} = softmax((Q^{ij}W_{q}^{ij})(K^{ij}W_{k}^{ij})^{T} / \sqrt(d_{e}))
\end{align}
where $A^{ij} \in \mathbb{R}^{n \times n}$. \attn{} applies a point-wise scaling factor to each element in the attention matrix before the softmax operation. These scaling factors are learned during fine-tuning on the task's support set $\mathcal{S}$. In particular, we define a learnable scaling tensor $A_{\alpha} \in \mathbb{R}^{n \times n \times L \times n_{h}}$. $A_{\alpha}$ can be reshaped as $\{ A_{\alpha}^{i} \}_{i=1}^{L}$ where $A_{\alpha}^{i} \in \mathbb{R}^{n \times n \times n_{h}}$ is the scaling tensor for each $i^{th}$ layer. For each attention head $j\in [1,n_{h}]$, the scaling matrix is defined as $A_{\alpha}^{ij} \in \mathbb{R}^{n \times n}$. 
\begin{align}
    A^{ij} = softmax(A_{\alpha}^{ij} \odot (Q^{ij}W_{q}^{ij})(K^{ij}W_{k}^{ij})^{T} / \sqrt(d_{e}))
\end{align}
During few-shot adaptation, only $A_{\alpha}^{ij}$ is learned along with the parameters in the DRA module from eTT. Note, $\{ W_{q}^{ij}, W_{k}^{ij}, W_{v}^{ij} \}$ are kept frozen for each $i^{th}$ layer and $j^{th}$ attention head. In principle, the scaling factor $A_{\alpha}$ replaces the attentive-prefix tuning (APT) module in eTT. This APT module uses $\sim\!$ 9$\%$ model parameters, whereas \attn{} uses only $\sim\!\! 1.2\%$ but still gives improved \mdshort{} performance. 
 \begin{figure}
    \hskip -0.3cm
    \vspace{-0.1cm}
  \includegraphics[width=9.2cm, height=4.2cm]
  {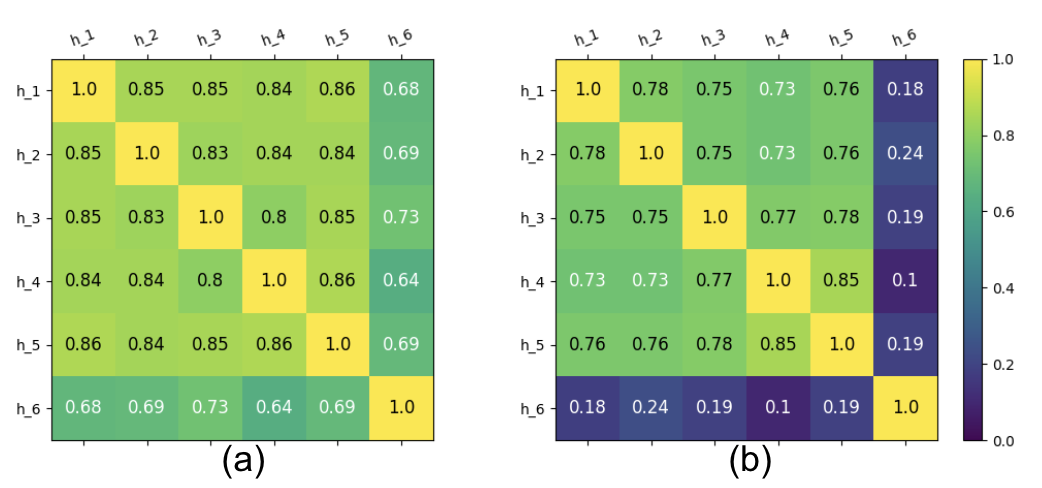}
 \caption{\label{attentionheatmap} \textbf{Different attention heads encode similar attention maps in self-supervised ViTs} -- (a) ViT-S/16(DINO); (b) ViT-S/16(DeiT). We compute the Pearson correlation between the attention scores of different heads: $h_{\_}i,\ \hspace{0.1cm} \forall i \in [1,n_{h}]$. Self-supervised ViTs encode attention across different heads more similarly than supervised ViTs. Correlation is averaged across examples from 100 tasks from each of the 10 domains in \mdshort{}.} 
 \vspace{-0.3cm}
\end{figure}

We also propose a light-weight extension of \attn{}, called \attnlite{}, which learns the same scaling parameters across {\it all} $n_{h}$ attention heads in a given layer, rather than different ones for each head. This is motivated by an observation that all $n_{h}$ attention heads in a layer have similar attention maps. We show this in~\Cref{attentionheatmap} where we plot the pairwise Pearson correlation~\cite{benesty2009pearson} between the attention values of different heads. Here, for self-supervised ViTs, we see strong correlation values between different heads in a given layer indicating that different heads encode similar kinds of attention maps. This is similar for supervised ViTs, however, the correlation values are slightly lower. Formally, for \attnlite{}, we define the scaling parameter for the $i^{th}$ layer as $A_{\alpha}^{i} \in \mathbb{R}^{n \times n}$ and $A_{\alpha}^{ij} = A_{\alpha}^{i}, \hspace{0.2cm} \forall j \in [1,n_{h}]$. \attnlite{} requires only 0.25$\%$ of the total parameters for ViT-S/16 and only 0.09$\%$ for ViT-B/16 which makes it an extremely light-weight module. 
In~\Cref{empirical}, we provide fine-grained results on the efficacy of both \attn{} and \attnlite{} for downstream few-shot adaptation. We also provide a PyTorch-like implementation of \attn{} and \attnlite{} in the Appendix. 

% In~\Cref{empirical}, we show that \attn{} results in the SoTA performance on \mdshort{} for self-supervised DINO pre-trained ViT-S/16 and is particularly beneficial to use under large domain shifts.

\begin{table*}[t!]
\hskip 0.1cm
\scalebox{0.80}{
\begin{tabular}{SSSSSSSSSSSS} \toprule
    {PEFT} & {MSCOCO} & {{\color{cyan}Traffic-Sign}} & {{\color{cyan} Omniglot}} & {Aircraft} & {DTD} & {VGG-Flower} & {{\color{cyan} Quickdraw}} & {Cu-birds} & {Fungi} & {Overall} & {Rank}\\ \midrule
    Full  & 61.5 & 87.3 & 78.7 & 75.4 & {\color{blue}86.9} & 94.2 & 73.6 & 85.4 & 54.7 & 77.5 & 6 \\ \midrule 
    \text{{\color{red} Adapter}} & 55.8  & 52.2 & 54.7  & 60.01 & 83.8 & 94.6  & 60.5 & 84.8 & 55.9 & 66.8 & 9 \\
    \text{{\color{red}Bias}}  & {\color{brown}63.4}  & {\color{brown}90.4} & 80.4  & 77.5 & 84.7 & 95.1  & {\color{brown}74.3} & 85.6 & 58.9 & 78.8 & 4\\
    \text{{\color{red}LoRA}} & 62.1  & 88.1 & {\color{brown}80.8}  & {\color{blue} 80.8} & {\color{brown}86.8}  & 94.8   & 72.7 & 85.8  & {\color{brown}59.8} & 78.9 & 3 \\
    \text{{\color{red}Ladder}}  & 55.7  & 52.2 & 54.7  & 60.01 & 83.8 & 94.6  & 60.5 & 84.8 & 55.9 & 67.0 & 8\\ 
    \text{{\color{red}Prompt-Shallow}}  & 52.7  & 58.9 & 61.8  &  62.9 & 83.0 & 94.2 & 66.0 & 83.4 & 55.5 & 68.7 & 7\\
    \text{{\color{red}Prompt-Deep}}  & 62.8  & 85.6 & 77.0  & 73.3 & 85.3 & {\color{brown}96.2} & 73.2  & {\color{brown}86.1} & 58.2 & 77.5 & 6\\
    \text{{\color{red}eTT}}  & 61.5  & 89.1 & 78.9  & 75.8 & 85.1 & 95.1  & 73.5  & {\color{brown}86.1} & 58.2 & 78.1 & 5 \\ \midrule
    {\color{magenta} \lnorm{}}  & {\color{blue}64.2}  & 91.2 & 77.9  & 75.3 & 84.4 & {\color{blue}96.9}  & {\color{blue}74.7} & {\color{blue}87.5} & {\color{blue}59.9} & 79.1 & \textbf{2}\\
    {\color{magenta} \attn{}} & 61.9  &  {\color{blue}91.4} & {\color{blue}80.9}  & 78.8 & 85.8 & 95.9  & 74.4 & 86.7 & 59.01 & {\color{blue} 79.4} & \textbf{1}\\
    {\color{magenta} \attnlite{}} & 61.6  &  91.0 & 80.2  & 77.9 & 85.8 & 96.0  & 73.9 & 86.7 & 59.0 & 79.1 &  \textbf{2}\\ \bottomrule
\end{tabular}}
\caption{\label{table_fewshot} \textbf{Our strong baselines, \lnorm{} and \attn{}, rank in the top 2 of all PEFT methods on the few-shot classification benchmark, \md{}.} Results shown for a ViT-S/16 (DINO), and exclude the ImageNet split. {\color{blue} Blue: Best overall performing PEFT method}; {\color{red}Red: existing PEFT methods.}; {\color{brown}Brown: Best performing module amongst existing PEFT methods.}; {\color{magenta}Magenta: Strong baselines proposed in our paper.}; {\color{cyan} Domains with large domain shifts from ImageNet-1k.} }
\vspace{-0.2cm}
\end{table*}
% \textbf{Note on one failure case.} We found that placing the scaling parameters $A_{\alpha}$ after the softmax operation in each layer and performing a subsequent softmax operation to normalize the attention weights performed significantly worse than inserting the scaling parameters before the softmax (see Sec.(2) in Appendix). 
\section{Empirical Results on \md{}}
\label{empirical}
We use our wide-scale empirical study to derive novel insights on PEFT methods for few-shot classification. In particular, we use our results on \mdshort{} to answer the following key questions: 
\circled{1} Do PEFT methods rank similarly across different pre-training architectures and learning objectives? 
\circled{2} How does the fine-tuning algorithm influence the performance of a PEFT method?  
\circled{3} Is the optimal PEFT method different for different data domains? 
\circled{4} Can PEFT modules be dropped from certain positions in the feature extractor? This can lead to significant memory and storage savings during few-shot deployment.
\circled{5} What is the impact of PEFT methods on the downstream fine-tuning run-time? These are critical factors when deploying a few-shot classifier in the wild. We also show that our two simple but {\it strong} baselines, \lnorm{} and \attn{}, perform better than full fine-tuning and all top-performing PEFT methods.
\vspace{-0.1cm}
\subsection{Consistency Across Pre-Training Models}
\label{ref_pretrain}
We analyse the influence of pre-training model by ranking the performance of different PEFT methods across the different pre-training objectives and architectures described in~\Cref{exp_test_bed}. To isolate the role of the pre-trained model, for each run, we keep all other variables constant including the fine-tuning algorithm, position of the modules, and hyperparameters. We report the results using the \protoaug{} fine-tuning algorithm in~\Cref{pretrain_consistency}, and include results for \ncc{} and \linear{} in the Appendix. 
\begin{figure}
 \vspace{-0.3cm}
    \hskip 0.2cm
  \includegraphics[width=8.0cm, height=6.0cm]
  {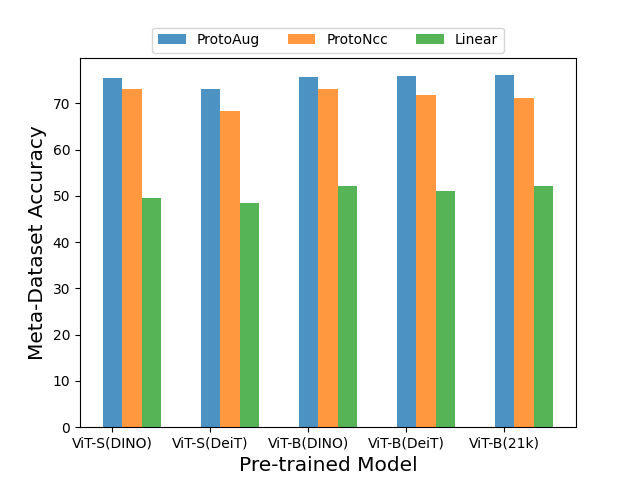}
 \caption{\label{FTStrategy} \textbf{With PEFT methods, we find \protoaug{} to have the best performance on \md{}, while \linear{} performs the worst. }\mdshort{} accuracy averaged over all 10 PEFT methods with different fine-tuning algorithms.}
 \vspace{-0.2cm}
\end{figure}

\begin{figure}
 \vspace{-0.4cm}
    \hskip 0.65cm
  \includegraphics[width=7.0cm, height=4.5cm]
  {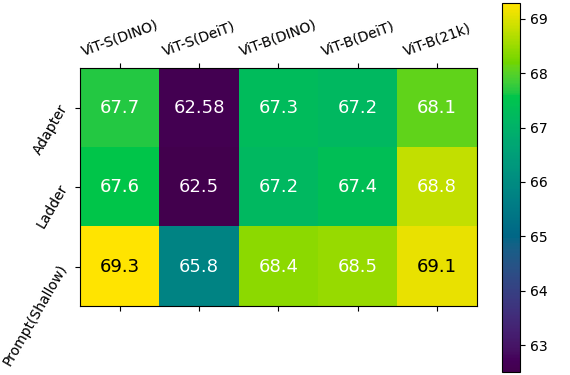}
  \vspace{-0.2cm}
 \caption{\label{worse} \textbf{Adapters, ladder tuning and shallow prompt-tuning are the worst-performing PEFT methods on \md{} (averaged across all the domains in \md{}) showing a significant drop of ($\sim10\%$) compared to other methods. }}
 \vspace{-0.5cm}
\end{figure}

\textbf{Existing PEFT methods. } In~\Cref{pretrain_consistency}-(a), we find that PEFT methods rank inconsistently, with no single best approach, across the different pre-trained models. In~\Cref{pretrain_consistency}-(b), we plot the Spearman correlation of the PEFT method's ranking between different pre-trained models. We observe that the correlation values across all pairs of pre-trained models are not consistently high, suggesting that existing PEFT methods do {\it not} generalize similarly for different pre-trained architectures and objectives. We also find that adapters, ladder-tuning and shallow prompt-tuning all have sub-par performances on \mdshort{} ($\sim\!10\%$ drop) when compared to LoRA, bias-tuning, eTT and deep prompt-tuning (see \Cref{worse}). We also highlight that shallow prompt-tuning struggles with few-shot classification on \mdshort{} despite performing competitively on transfer learning natural tasks in VTAB~\cite{vtab_}. 
%  \begin{figure}
%  \vspace{-0.4cm}
%     \hskip 0.65cm
%   \includegraphics[width=7.0cm, height=4.5cm]
%   {worse.png}
%   \vspace{-0.2cm}
%  \caption{\label{worse} \textbf{Adapters, ladder tuning and shallow prompt-tuning are the worst-performing PEFT methods on \md{} (averaged across all the domains in \md{}) showing a significant drop of ($\sim10\%$) compared to other methods. }}
%  \vspace{-0.8cm}
% \end{figure}
Deep prompt-tuning~\cite{vtab_}, which is the state-of-the-art PEFT module on VTAB, performs competitively on \mdshort{} across all pre-trained models, but falls short of methods like eTT~\cite{xu2022exploring}, LoRA~\cite{lora}, bias-tuning~\cite{bitfit} and full model-tuning~\cite{shell_simple_baselines} 
 (see~\Cref{pretrain_consistency}). This result highlights that strongly performing PEFT methods for transfer learning {\it do not} generalize well to the challenging few-shot setting of \mdshort{}. eTT~\cite{xu2022exploring} for ViT-S/16(DINO) outperforms full model-tuning, but also lags behind LoRA and bias-tuning. Overall, we find bias-tuning~\cite{bitfit} to consistently rank amongst the top 4 across all the pre-training models, outperforming many of the more complex PEFT methods.
\textbf{Our strong baselines. } From~\Cref{pretrain_consistency}, we find that our strong baselines, \lnorm{} and \attn{}, perform strongly across all the pre-trained models on~\mdshort{}. In particular, \lnorm{} performs the best for supervised ViTs (pre-trained on ImageNet-1k and ImageNet-21k) consistently. We also highlight that for supervised ViTs, none of the PEFT methods except \lnorm{} reaches performance close to full fine-tuning. \attn{}, which is around 9x more parameter-efficient than eTT, has the best few-shot performance for self-supervised ViTs pre-trained using DINO~\cite{dino}. For self-supervised ViTs, \lnorm{} performs closely to \attn{} and ranks in the top 2 methods.  

% These results set fair baselines which can be used as a reference to {\it surgically} design PEFT methods for different pre-training paradigms in the context of few-shot classification. 
%  \begin{figure}
%  \vspace{-0.1cm}
%     \hskip 0.2cm
%   \includegraphics[width=8.0cm, height=6.0cm]
%   {finetune.png}
%  \caption{\label{FTStrategy} \textbf{\mdshort{} accuracy averaged over all 10 PEFT methods with different fine-tuning algorithms}. With PEFT methods, we find \protoaug{} to have the best performance on \md{}, with \linear{} having sub-par performance.}
%  \vspace{-0.5cm}
% \end{figure}
\subsection{Effect of Fine-tuning Algorithm}
\label{finetuningstrategy}
We quantify the impact of 3 different algorithms for fine-tuning the parameters in PEFTs: \linear{}, \protoaug{} and \ncc{}. We find that \protoaug{} outperforms \ncc{} and strongly outperforms \linear{} across all pre-training objectives and PEFT methods including full model tuning (\Cref{FTStrategy}). In some cases, \protoaug{} and \ncc{} outperform \linear{} by as much as $20\%$. We also find that for self-supervised pre-training objectives like DINO~\cite{dino}, the gap between \protoaug{} and \ncc{} is $\sim\!\!\!2.2\%$, whereas for supervised objectives like DeiT~\cite{deit} this gap is higher at $\sim\!\!\!4.7\%$ (for both ImageNet-1k and ImageNet-21k initializations). Since the only difference between \protoaug{} and \ncc{} is that the query set is an augmented version of the support set, this suggests that applying augmentations during few-shot (meta) fine-tuning is more effective with supervised than self-supervised objectives. We also note that when using full model fine-tuning, \protoaug{} outperforms \ncc{} by $\sim\!\!5\%$ for DINO and by $\sim\!\!6.7\%$ for DeiT objectives. This gap is higher than when used with other PEFT methods (see~\Cref{ft_gap}). This suggests that \protoaug{}'s efficacy decreases when used in conjunction with PEFT methods.
\begin{table}[H]
\vspace{-0.2cm}
\hskip 0.0cm
\scalebox{0.8}{
\begin{tabular}{SSSS} \toprule
    {Method} & {\protoaug{}} & {\ncc{}} & {Performance Gap}\\ \midrule
    \text{Full Tuning (DINO)} & 77.2 & 72.2 & \textbf{$\Delta$ 5.0$\%$}\\
     \text{All PEFTs (DINO)} & 75.4 & 73.2 & $\Delta$ 2.2$\%$ \\  \midrule
    \text{Full Tuning (DeiT)} & 78.1 & 71.38 & \textbf{$\Delta$ 6.7$\%$} \\
    \text{All PEFTs (DeiT)} & 73.1 & 68.4 & $\Delta$4.7$\%$ \\
  \bottomrule
\end{tabular}}
\caption{\label{ft_gap} \textbf{The performance gap between \protoaug{} and \ncc{} is more with full fine-tuning than when used with PEFT methods.} }
\end{table}

\subsection{Comparing Performance Across Domains}
\label{consistencydomains}
\vspace{-0.1cm}
We leverage the distinct sub-datasets in \mdshort{} to compare the performance of PEFT methods across domains. Since each sub-datasets has a different degree of domains shifts from the pre-training dataset (ImageNet), we also evaluate the robustness of different PEFT methods to these shifts. In~\Cref{table_fewshot}, we show these results with a ViT-S/16 pre-trained with DINO, and observe that none of the PEFT methods are consistently the best across domains. We show similar results for other pre-trained ViTs in the Appendix.

\textbf{Existing PEFT methods. } We observe that deep prompt-tuning is the best PEFT method for domains with smaller degrees of shift from ImageNet such as Cu-Birds and VGG-Flower. It is second best on MS-COCO, which is also similar to ImageNet. We find, however, that for larger domain shifts such as Omniglot, Quickdraw and Traffic-Sign it struggles, with LoRA and bias-tuning showing stronger performance. This is similarly the case for adapters, LoRA, and ladder-tuning which also perform poorly on larger domain shifts and have the lowest average performance on \mdshort{} generally. 
% Next, we analyse the PEFT methods such as adapters, prompt-tuning(shallow) and ladder-tuning, which have the lowest average performances on \mdshort{}. We specifically find that these methods are not able to effectively adapt to domains where there is a significant domain shift such as Traffic-Sign, Omniglot, and Quickdraw. 

\textbf{Our strong baselines. } We find that \lnorm{} in~\Cref{table_fewshot} outperforms all existing PEFT methods in 5 out of the 9 domains, with \attn{} lagging behind it only slightly in these 5 domains. However, for domains with a larger shift (e.g., Omniglot, Traffic-Sign), \attn{} performs better than \lnorm{}. Even for Quickdraw, where there is a significant shift, \attn{} and \lnorm{} perform almost similarly. Overall on \mdshort{}, \attn{} ranks the best in terms of few-shot performance. These results suggest that our two strong baselines can be used complementarily: when the domain shift from the pre-training dataset is high, \attn{} is better suited, whereas when the domain shift is low, \lnorm{} is the stronger approach. 
\begin{table*}[t!]
\hskip 0.1cm
\scalebox{0.75}{
\begin{tabular}{SSSSSSSSSSSS} \toprule
    {Model} & {Full} & {{\color{red}Adapter}} & {{\color{red}Bias}} & {{\color{red}LoRA}} & {{\color{red}Ladder}} & {{\color{red}Prompt-Deep}} & {{\color{red}Prompt-Shallow}} & {{\color{red}eTT}} & {{\color{magenta}\lnorm{}}} &{{\color{magenta}\attn{}}} & {{\color{magenta}\attnlite{}}}\\ \midrule
    \text{ViT-S(DINO)}  & 63.1 & 62.6 &{\color{brown}67.1} & 66.4 & 62.7 & 65.7 & 51.8 & 65.6 & {\color{blue}67.8} & 67.2 & 66.9\\
    \text{ViT-S(DeiT)} & 66.6  & 66.8 & 66.4  & 67.6 & 66.9 & 66.7  & 63.4 & {\color{brown}68.4} & {\color{blue}68.8} & 67.1 & 66.2\\ 
    % \text{ViT-B(DINO)} & x  & x & x  & x & x & x  & x & x & x & x & x\\ 
    % \text{ViT-B(DeiT)} & x  & x & x  & x & x & x  & x & x & x & x & x\\ 
    % \text{ViT-B(21k)} & x  & x & x  & x & x & x  & x & x & x & x & x\\ 
    \bottomrule
\end{tabular}}
\caption{\label{table_fewshot_orbit} \textbf{\lnorm{} results in the best performance on \orbit{} while \attn{} is extremely competitive.} Frame accuracy results are shown for a ViT-S/16. {\color{red} Red: Existing PEFT methods;} {\color{magenta} Magenta: Strong baselines proposed in our paper.}; {\color{brown}Brown: Best performing module amongst existing PEFT methods.}; {\color{blue} Best overall performing PEFT method}; We provide additional results with ViT-B/16 in the Appendix. }
\vspace{-0.3cm}
\end{table*}
Our results highlight that current PEFT methods are not robust to varying degree of domain shifts and requires rethinking the current designs of PEFT modules to be uniformly robust to all domain shifts. Overall, our proposed strong baselines lead to the best performance in 7 out of 9 domains. 
% and emphasize that there is a need to {\it surgically} design PEFT methods for domains with different degrees of distribution shifts. Overall, the strong baselines proposed in our paper lead to the best performance in 7 out of 9 downstream domains.  

\textbf{Performance of \attnlite{}}. \label{attn_lite_description} We observe from~\Cref{table_fewshot} that \attnlite{} performs similarly to \lnorm{} but slightly worse than \attn{} (by around $0.5-0.7\%$) on larger domain shifts for self-supervised ViT-S/16(DINO). For smaller domain shifts, \attnlite{} matches the performance of \attn{}. For supervised ViTs, we find that \attnlite{} lags behind \attn{} by a larger margin of $1.2-1.8\%$ for large domain shifts (see Appendix for results). The decrease in the effectiveness of \attnlite{} for supervised ViTs can be attributed to the fact, that different heads encode attention maps less similarly than self-supervised ViTs. Therefore, learning a separate set of scaling parameters for different heads is more beneficial for few-shot adaptation. 
%  \begin{figure}
%     \hskip 0.6cm
%   \includegraphics[width=7.5cm, height=5.4cm]
%   {layer_drop.png}
%  \caption{\label{layerdrop} \textbf{ Dropping \lnorm{} from earlier layers in the ViT for large domain shifts (e.g., Traffic-Sign, Quickdraw, Omniglot) leads to a large drop in accuracy.} We investigate the effect of inserting \lnorm{} at different layers in the ViT.}
%  \vspace{-0.3cm}
% \end{figure}

\subsection{Can we drop PEFT modules from ViT layers? }
\label{ref_drop}
\begin{figure}
    \hskip 0.6cm
  \includegraphics[width=7.5cm, height=5.4cm]
  {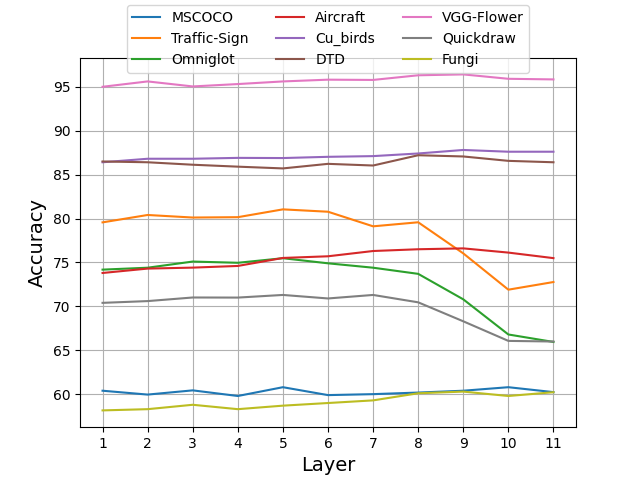}
 \caption{\label{layerdrop} \textbf{ Dropping \lnorm{} from earlier layers in the ViT for large domain shifts (e.g., Traffic-Sign, Quickdraw, Omniglot) leads to a large drop in accuracy.} We investigate the effect of inserting \lnorm{} at different layers in the ViT.}
 \vspace{-0.3cm}
\end{figure}

In Secs.~\ref{consistencydomains} and \ref{finetuningstrategy}, the PEFT modules are inserted in each of the 12 layers of the ViT. In this section, we use our strong baselines, \lnorm{} and \attn{}, to examine if dropping PEFT modules from the majority of layers impacts performance. Specifically, we insert a PEFT module in the final layer of the ViT and another in 1 other layer (between 1-11). We vary the position of the second PEFT and observe its impact on performance (\Cref{layerdrop}).  

\textbf{Results.} From~\Cref{layerdrop}, we find that inserting the PEFT into the later layers of the ViT improves the performance more than inserting it in the earlier layers for domains with a small degree of shift from ImageNet (e.g., MSCOCO, DTD, VGG-Flower, Cu$\_$birds). However, for large domain shifts such as in Traffic-Sign, Quickdraw and Omniglot, we find that inserting \lnorm{} in the earlier layers is crucial. In particular for these domains, we find that inserting \lnorm{} {\it only} in the later layers results in $\sim\!\!10\%$ drop in accuracy . We observe similar results for \attn{} (see Appendix). 
%  \begin{figure}
%     \hskip 0.6cm
%   \includegraphics[width=7.5cm, height=5.4cm]
%   {layer_drop.png}
%  \caption{\label{layerdrop} \textbf{ Dropping \lnorm{} from earlier layers in the ViT for large domain shifts (e.g., Traffic-Sign, Quickdraw, Omniglot) leads to a large drop in accuracy.} We investigate the effect of inserting \lnorm{} at different layers in the ViT.}
%  \vspace{-0.3cm}
% \end{figure}
% \subsection{Impact on Calibration}
% In this section, we investigate the effectiveness of PEFT methods on calibration error. Calibration error is an important safety metric when deploying deep learning models, however its interplay with PEFT methods is not well understood. From Fig. (xx), we find that ... {\color{blue} TODO}
\subsection{Impact on Fine-tuning Time}
Although PEFT methods save a significant amount of storage, they are not necessarily faster to fine-tune compared to full model fine-tuning as the modules are often deep inside the network and gradients must be back-propagated through the entire backbone. We empirically quantify this by measuring the fine-tuning time per task across all \mdshort{} domains. In particular, we compute the speedup factor of the PEFT methods when compared to full fine-tuning. We find that all the PEFT methods, except ladder-tuning, provide a speedup of {\it only} 1.3-1.9x compared to full fine-tuning. Ladder-tuning, since it does not require any gradient computation through the backbone, has a greater speedup of 3.3x compared to full fine-tuning (See Appendix).

\vspace{-0.1cm}
\section{Results on Tasks from \orbit{}}
In this section, we compare PEFT methods on the challenging personalization tasks from \orbit{}. We modify the task sampling procedure in \orbit{} to decrease the maximum size of the support set sizes, so that \protoaug{} can be used for fine-tuning. We provide the detailed task sampling procedure in the Appendix.
% \textbf{Experimental Setup for \orbit{}.} Fine-tuning on tasks with \protoaug{} from \orbit{}'s default sampler is memory-intensive as the support set sizes are often greater than 1000 frames. We, therefore, modify the task sampling procedure so that fine-tuning can be run on an A6000 GPU with 48GB memory: (i) For each video, we use the random sampling configuration and change the \texttt{clip\_cap} parameter from 200 to 50. (ii) We change the \texttt{context\_shot} parameter from 15 to 3 to ensure that per object class, a maximum number of 3 videos are sampled in the support set. (iii) We resize the frames to 128$\times$128 from 224$\times$224.  These changes enable us to fine-tune ViT-S/16 and ViT-B/16 architectures using \protoaug{} described in~\ref{finetuningstrategy}.  In total, we sample the same number of tasks - 50 tasks per test user - giving 850 tasks in total. We report the average frame accuracy over all these tasks. 

\textbf{Overall Results. } From~\Cref{table_fewshot_orbit}, we find that bias-tuning and eTT have the best performances amongst the existing PEFT methods for ViT-S/16 (DINO) and ViT-S/16 (DeiT), respectively. These results reinforce our previous finding that different PEFT methods may be suited to different pre-training objectives. Overall, we find that \lnorm{} results in the best few-shot performance for both self-supervised (DINO) and supervised (DeiT) pre-training objectives across all PEFT methods. \attn{} ranks in the top 2 for DINO, however, for DeiT we find its performance slightly drops but still ranks within the top 4 PEFT methods. 
% For \orbit{}, we want to highlight that in order to take full advantage of the original sampling configurations -- the fine-tuning strategies from~\ref{exp_test_bed} can be integrated with the LITE framework~\cite{lite_ml}.  

\section{Conclusion}
In this paper, we perform a large-scale controlled empirical study of a wide range of top-performing PEFT methods across large-scale challenging few-shot classification benchmarks such as \mdshort{} and \orbit{}. Our main finding is that two embarrassingly simple approaches -- \lnorm{} and \attn{} -- beat all the PEFTs we evaluated and set new state-of-the-art results on \mdshort{} and competitive results on \orbit{}. Our proposed strong baselines are easy-to-implement, significantly less complex and parameter-intensive. The large scale of our empirical study also uncovers several novel empirical insights, including that there is no one-size-fits-all PEFT method across different pre-training architectures, objectives (self-supervised or supervised), pre-training datasets (ImageNet-1k or ImageNet-21k) and downstream domains with different degrees of distribution shifts. Together, our experimentally consistent suite of experiments and {\it strong} baselines supports the future study of parameter-efficient fine-tuning approaches for few-shot classification, but calls for rethinking current practices in light of simple but effective baselines. 
{\small
\bibliographystyle{ieee}
\bibliography{egbib}
}

\appendix 

\section{Pre-trained Models}
In our experimental setup, we use 5 distinct pre-trained models: (i) ViT-S/16 pre-trained using DINO~\cite{dino}; (ii) ViT-B/16 pre-trained using DINO~\cite{dino}; (iii) ViT-S/16 pre-trained using DeiT~\cite{deit}; (iv) ViT-B/16 pre-trained using DeiT~\cite{deit} and (v) ViT-B/16 pre-trained using a supervised learning objective similar to DeiT.  (i, ii, iii, iv) use ImageNet-1k as the pre-training dataset, while (v) uses ImageNet-21k as the pre-training dataset. The use of DINO as a pre-training strategy is motivated by~\cite{dino} where the authors propose using ViT-S/16(DINO) as a strong baseline for few-shot classification. ViT-S/16 and ViT-B/16 pre-trained with DeiT is chosen to understand if the methods developed for self-supervised DINO (e.g., eTT~\cite{xu2022exploring}) can be generalized to supervised pre-training strategies. ViT-B/16 pre-trained with ImageNet-21k is chosen to understand the effectiveness of distinct PEFT methods for large-scale pre-training datasets beyond ImageNet-1k. We also highlight that existing PEFT methods are developed only in the light of one pre-trained model. For e.g., eTT~\cite{xu2022exploring} is designed for ViT-S/16(DINO) whereas prompt-tuning~\cite{vtab_} is designed for ViT-B/16 pre-trained with ImageNet-21k. In comparison, our large-scale study encompasses : (i) self-supervised models; (ii) supervised models; (iii) models pre-trained on large-scale datasets. 
\section{Description of Existing PEFT Methods}
\textbf{Adapters.} We use the improved adapter design from~\cite{adapter_pffeifer} where the adapter block is inserted {\it only} once in each layer. We use GeLU activation for the adapter layer and set the hyper-parameter reduction$\_$dim as 8, which is the default configuration.  The adapter is initialized with a zero-initialization.

\textbf{Prompt-Shallow.} We use the shallow prompt-tuning design from~\cite{vtab_} and use 8 as the length of the prompt. In Prompt-Shallow, the learnable prompts are only inserted with the input before the first block in the ViT.  

\textbf{Prompt-Deep.} For deep prompt-tuning, learnable prompts of length 8 are inserted in each layer of the ViT. At each layer, the embeddings of the prompts from the earlier layers are discarded and the new learnable prompts are inserted. This ensures that the input to each layer in the ViT is fixed to be the total number of original tokens and prompt tokens, together. 

\textbf{eTT.} We use the default hyperparameter configurations from~\cite{xu2022exploring} for eTT. For each task, the number of prefixes to be attached to the key and value matrices is dynamic and is set to be the number of classes in the sampled task. eTT uses \ncc{} during fine-tuning, whereas we use both \protoaug{} and \ncc{}. We highlight that while eTT~\cite{xu2022exploring} use a prototypical-regularizer -- we do not use any external regularization to ensure a fair comparison to other PEFT methods which do not use prototypical-regularizer, but use \protoaug{} or \ncc{} during fine-tuning. 

\textbf{Ladder.} In ladder-tuning~\cite{ladder} -- adapter like blocks are attached in a ladder like structure, where each block takes input from the previous ladder block and the corresponding block in the pre-trained ViT. We set the parameter reduction$\_$dim=8 in each of the ladder block. 

\textbf{Bias.} In bias-tuning, the bias parameters of each and every block in the ViT is updated. 

\textbf{LoRA.} We set the rank of LoRA blocks to be 8 across our entire experimental setup. In our experimental setup, we follow~\cite{lora} and restrict our study to only performing the low-rank decomposition of all the projection matrices in the attention blocks. 

\section{Task Sampling Details for \mdshort{}}
We follow the original task sampling procedure from~\cite{metadataset}, where 600 tasks per domain are sampled. Following~\cite{shell_simple_baselines}, we resize the images to 128x128 during fine-tuning. We do not use any augmentations except for \protoaug{} fine-tuning to generate the query set. For downstream domains, we do use the ImageNet split, as the pre-trained models use ImageNet-1k or ImageNet-21k as the pre-training dataset. 
\section{Task Sampling Details for \orbit{}}
Fine-tuning on tasks with \protoaug{} and \ncc{} from \orbit{}'s default sampler is memory-intensive as the support set sizes are often greater than 1000 frames (images). We, therefore, modify the task sampling procedure so that the fine-tuning procedure can be run on an A6000 GPU with 48GB memory: (i) For each video, we use the random sampling configuration and change the \texttt{clip\_cap} parameter from 200 to 50. (ii) We change the \texttt{context\_shot} parameter from 15 to 3 to ensure that per object class, a maximum number of 3 videos are sampled in the support set. (iii) We resize the frames to 128$\times$128 from 224$\times$224.  These changes enable us to fine-tune ViT-S/16 and ViT-B/16 architectures using \protoaug{} or \ncc{}.  In total, we sample the same number of tasks - 50 tasks per test user - giving 850 tasks in total. We report the average frame accuracy over all these tasks. The main disadvantage of \protoaug{} or \ncc{} is the issue of creating two computational graphs during the fine-tuning procedure -- one for the support set and another for the query set. This can be computationally expensive for fine-tuning large models such as ViT-B/16. In light of this, specifically for ViT-B/16, we further integrate the fine-tuning procedure with LITE~\cite{lite_meml}. 

\section{Additional Results on \mdshort{}}
We provide additional composite results on \mdshort{} in~\Cref{table_fewshot_ViT_B} for ViT-B/16(DINO),~\Cref{table_fewshot_ViT_s_deit} for ViT-S/16(DeiT), ~\Cref{table_fewshot_ViT_b_deit} for ViT-B/16(DeiT) and ~\Cref{table_fewshot_ViT_b_21k} for ViT-B/16(ImageNet-21k).  For self-supervised ViT-B/16(DINO), we find that \attn{} leads to the best few-shot performance. For supervised ViTs pre-trained on both ImageNet-1k and ImageNet-21k, we find that \lnorm{} leads to the best few-shot performance, while having the most parameter efficient PEFT method.  
\begin{table*}[t!]
\hskip 0.1cm
\scalebox{0.80}{
\begin{tabular}{SSSSSSSSSSSS} \toprule
    {PEFT} & {MSCOCO} & {{\color{cyan}Traffic-Sign}} & {{\color{cyan} Omniglot}} & {Aircraft} & {DTD} & {VGG-Flower} & {{\color{cyan} Quickdraw}} & {Cu-birds} & {Fungi} & {Overall} & {Rank}\\ \midrule
    Full  & 64.1 & 88.9 & 80.0 & 78.9 & 85.9 & 94.8 & 73.4 & 85.0 & 60.0 & 79.0 & 5 \\ \midrule 
    \text{{\color{red} Adapter}} & 58.0  & 50.4 & 60.2  & 53.7 & 84.0 & 94.9  & 59.7 & 79.4 & 56.0 & 66.3 & 8 \\
    \text{{\color{red}Bias}}  & 64.6  & 90.7 & 80.6  & 78.2 & 86.8 & 95.7  & 74.9 & 84.3 & 58.1 & 79.3 & 3\\
    \text{{\color{red}LoRA}} & 64.1  & 89.2 & 81.0  & 80.8 & 85.8  & 94.8   & 71.7 & 86.7  & 59.4 & 79.2 & 4 \\
    \text{{\color{red}Ladder}}  & 57.9  & 50.3 & 60.2  & 53.7 & 84.0 & 94.9  & 59.7 & 79.4 & 55.8 & 66.2 & 9\\ 
    \text{{\color{red}Prompt-Shallow}}  & 58.2  & 57.7 & 60.8  &  58.4 & 83.9 & 94.9 & 60.5 & 78.7 & 55.5 & 67.6 & 7\\
    \text{{\color{red}Prompt-Deep}}  & 64.0  & 82.8 & 76.5  & 73.0 & 86.4 & 95.9 & 73.0  & 84.9 & 58.5 & 77.2 & 1\\
    \text{{\color{red}eTT}}  & 63.6  & 88.2 & 79.2  & 76.1 & 86.3 & 95.5  & 74.2  & 84.1 & 57.4 & 78.3 & 6 \\ \midrule
    {\color{magenta} \lnorm{}}  & 65.2  & 91.1 & 80.1  & 80.9 & 87.5 & 94.9  & 74.1 & 86.1 & 59.0 & 79.8 & \textbf{2}\\
    {\color{magenta} \attn{}} & 64.4  &  91.2 & 81.02  & 79.9 & 86.9 & 95.5  & 74.6 & 86.1 & 59.8 & {\color{blue} 79.9} & \textbf{1}\\ \bottomrule
\end{tabular}}
\caption{\label{table_fewshot_ViT_B} \textbf{Composite results for ViT-B/16(DINO) on \mdshort{}: \attn{} has the best performance on \mdshort{}. }}
\vspace{-0.0cm}
\end{table*}

\begin{table*}[t!]
\hskip 0.1cm
\scalebox{0.80}{
\begin{tabular}{SSSSSSSSSSSS} \toprule
    {PEFT} & {MSCOCO} & {{\color{cyan}Traffic-Sign}} & {{\color{cyan} Omniglot}} & {Aircraft} & {DTD} & {VGG-Flower} & {{\color{cyan} Quickdraw}} & {Cu-birds} & {Fungi} & {Overall} & {Rank}\\ \midrule
    Full  & 63.9 & 86.9 & 79.2 & 74.8 & 84.6 & 93.9 & 73.4 & 83.1 & 54.0 & 77.0 & 3 \\ \midrule 
    \text{{\color{red} Adapter}} & 57.6  & 50.6 & 52.9  & 47.5 & 76.4 & 82.8  & 57.8 & 75.0 & 42.0 & 60.2 & 9 \\
    \text{{\color{red}Bias}}  & 63.4  & 83.7 & 78.5  & 72.2 & 84.8 & 93.6  & 72.0 & 83.5 & 55.4 & 76.3 & 4\\
    \text{{\color{red}LoRA}} & 63.3  & 87.3 & 80.4  & 77.2 & 84.5  & 93.9   & 71.5 & 83.2  & 52.9 & 77.1 & 2 \\
    \text{{\color{red}Ladder}}  & 57.6  & 50.5 & 52.9  & 47.6 & 76.4 & 82.8  & 57.8 & 75.0 & 42.0 & 60.3 & 8 \\ 
    \text{{\color{red}Prompt-Shallow}}  & 60.1  & 58.0 & 60.3  &  50.15 & 77.3 & 83.6 & 64.5 & 76.4 & 45.1 & 63.9 & 7\\
    \text{{\color{red}Prompt-Deep}}  & 63.7  & 83.2 & 74.2  & 70.1 & 83.7 & 93.7 & 71.5  & 81.9 & 51.9 & 74.9 & 6\\
    \text{{\color{red}eTT}}  & 64.4  & 84.0 & 72.3  & 70.0 & 84.2 & 93.8  & 71.9  & 82.7 & 54.7 & 75.3 & 5 \\ \midrule
    {\color{magenta} \lnorm{}}  & 64.1  & 86.4 & 79.3  & 73.9 & 84.8 & 94.1  & 73.1 & 83.5 & 55.0 & {\color{blue}77.2} & \bf{1}\\
    {\color{magenta} \attn{}} & 63.8  &  86.3 & 77.5  & 72.7 & 84.4 & 93.3  & 72.4 & 82.7 & 53.0 & 76.3 & 4\\ \bottomrule
\end{tabular}}
\caption{\label{table_fewshot_ViT_s_deit} \textbf{Composite results for ViT-S/16(DeiT) on \mdshort{}: \lnorm{} has the best performance on \mdshort{}.} }
\vspace{-0.0cm}
\end{table*}

\begin{table*}[t!]
\hskip 0.1cm
\scalebox{0.80}{
\begin{tabular}{SSSSSSSSSSSS} \toprule
    {PEFT} & {MSCOCO} & {{\color{cyan}Traffic-Sign}} & {{\color{cyan} Omniglot}} & {Aircraft} & {DTD} & {VGG-Flower} & {{\color{cyan} Quickdraw}} & {Cu-birds} & {Fungi} & {Overall} & {Rank}\\ \midrule
    Full  & 65.2 & 89.7 & 81.3 & 79.2 & 84.9 & 94.5 & 74.8 & 84.5 & 56.1 & 78.9 & 2 \\ \midrule 
    \text{{\color{red} Adapter}} & 58.1  & 54.5 & 61.7  & 53.7 & 79.5 & 88.6  & 61.2 & 78.8 & 46.5 & 64.7 & 10 \\
    \text{{\color{red}Bias}}  & 64.7  & 89.4 & 79.6  & 75.5 & 84.5 & 94.02  & 74.4 & 85.1 & 54.5 & 77.9 & 5\\
    \text{{\color{red}LoRA}} & 64.3  & 89.9 & 82.1  & 80.2 & 84.5  & 94.2   & 72.9 & 85.0  & 55.3 & 78.7 & 3 \\
    \text{{\color{red}Ladder}}  & 58.2  & 54.5 & 61.7  & 53.7 & 79.5 & 88.8  & 61.2 & 78.8 & 46.5 & 64.8 & 9 \\ 
    \text{{\color{red}Prompt-Shallow}}  & 59.5 & 57.2 & 64.5  &  54.9 & 79.7 & 88.5 & 78.4 & 65 & 47.55 & 66.1 & 8\\
    \text{{\color{red}Prompt-Deep}}  & 66.3  & 84.2 & 75.6  & 70.5 & 84.9 & 94.2 & 72.4  & 83.4 & 53.1 & 76.0 & 7\\
    \text{{\color{red}eTT}}  & 63.7  & 88.0 & 76.7  & 73.3 & 84.7 & 93.4  & 73.6  & 84.6 & 53.8 & 76.8 & 6 \\ \midrule
    {\color{magenta} \lnorm{}}  & 65.4  & 89.5 & 81.3  & 78.9 & 85.0 & 94.8  & 75.0 & 84.2 & 56.2 & {\color{blue}79.0} & \bf{1}\\
    {\color{magenta} \attn{}} & 63.9  &  88.2 & 79.6  & 77.2 & 84.8 & 94.1  & 74.6 & 84.6 & 54.6 & 78.0 & 4\\ \bottomrule
\end{tabular}}
\caption{\label{table_fewshot_ViT_b_deit} \textbf{Composite results for ViT-B/16(DeiT) on \mdshort{}: \lnorm{} has the best performance on \mdshort{}. } }
\vspace{-0.0cm}
\end{table*}

\begin{table*}[t!]
\hskip 0.1cm
\scalebox{0.80}{
\begin{tabular}{SSSSSSSSSSSS} \toprule
    {PEFT} & {MSCOCO} & {{\color{cyan}Traffic-Sign}} & {{\color{cyan} Omniglot}} & {Aircraft} & {DTD} & {VGG-Flower} & {{\color{cyan} Quickdraw}} & {Cu-birds} & {Fungi} & {Overall} & {Rank}\\ \midrule
    Full  & 67.6 & 90.2 & 74.7 & 77.9 & 83.5 & 99.0 & 74.2 & 93.6 & 59.6 & 80.0 & 2 \\ \midrule 
    \text{{\color{red} Adapter}} & 51.3  & 43.1 & 38.08  & 46.3 & 79 & 96.01  & 51.8 & 81.8 & 40.3 & 58.7 & 9 \\
    \text{{\color{red}Bias}}  & 66.1  & 88.5 & 61.3  & 73.7 & 84.9 & 98.8  & 73.2 & 91.2 & 59.0 & 77.4 & 5\\
    \text{{\color{red}LoRA}} & 65.6  & 89.2 & 63.0  & 77.7 & 84.7  & 98.7   & 72.9 & 91.8  & 57.9 & 77.9 & 3 \\
    \text{{\color{red}Ladder}}  & 51.9  & 43.1 & 38.08  & 46.7 & 79 & 96.01  & 51.8 & 81.9 & 40.3 & 58.9 & 8 \\ 
    \text{{\color{red}Prompt-Shallow}}  & 59.9 & 59.2 & 41.3  &  55.3 & 79.9 & 96.2 & 66.1 & 79.8 & 48.9 & 65.1 & 7\\
    \text{{\color{red}Prompt-Deep}}  & 66.0  & 86.1 & 66.8  & 70.6 & 84.8 & 98.6 & 72.4  & 89.8 & 55.1 & 76.6 & 6\\
    \text{{\color{red}eTT}}  & 64.3  & 88.1 & 64.1  & 71.3 & 83.5 & 98.1  & 71.8  & 92.1 & 56.7 & 76.6 & 6 \\ \midrule
    {\color{magenta} \lnorm{}}  & 67.5  & 90.05 & 73.4  & 77.6 & 86.3 & 99.0  & 74.0 & 93.6 & 59.8 & {\color{blue}80.2} & \bf{1}\\
    {\color{magenta} \attn{}} & 67.0  &  88.9 & 65.8  & 73.0 & 85.0 & 98.7  & 71.7 & 93.0 & 57.0 & 77.8 & 4\\ \bottomrule
\end{tabular}}
\caption{\label{table_fewshot_ViT_b_21k} \textbf{Composite results for ViT-B/16(ImageNet-21k) on \mdshort{}: \lnorm{} has the best performance on \mdshort{}.} }
\vspace{-0.0cm}
\end{table*}
\section{Further Results on Layer-wise Analysis}
\begin{figure}
    \hskip 0.6cm
  \includegraphics[width=7.5cm, height=5.4cm]
  {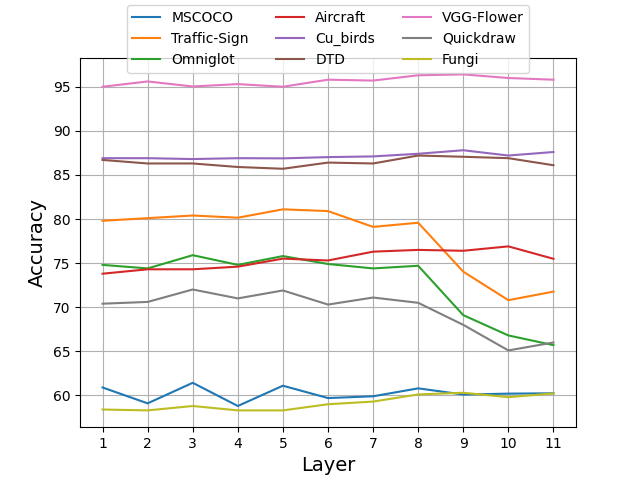}
 \caption{\label{layerdrop_attn} \textbf{ Similar to \lnorm{} -- dropping \attn{} from earlier layers in the ViT for large domain shifts (e.g., Traffic-Sign, Quickdraw, Omniglot) leads to a large drop in accuracy.} We investigate the effect of inserting \attn{} at different layers in the ViT.}
 \vspace{-0.1cm}
\end{figure}

In~\Cref{layerdrop_attn}, we find that similar to \lnorm{} -- dropping \attn{} from the earlier layers is detrimental to few-shot performance for downstream datasets, where there is a large domain shift (e.g., Quickdraw, Omniglot and Traffic-Sign) from the pre-training dataset (ImageNet-1k). 
\begin{table*}[t!]
\hskip 1cm
\scalebox{0.75}{
\begin{tabular}{SSSSSSSSSSSS} \toprule
    {Model} & {Full} & {{\color{red}Adapter}} & {{\color{red}Bias}} & {{\color{red}LoRA}} & {{\color{red}Ladder}} & {{\color{red}Prompt-Deep}} & {{\color{red}Prompt-Shallow}} & {{\color{red}eTT}} & {{\color{magenta}\lnorm{}}} &{{\color{magenta}\attn{}}} \\ \midrule
    \text{ViT-B/16(DINO)}  & 63.8 & 62.1 & 66.8 & 66.3 & 62.4 & 64.0 & 58.1 & 60.8 & {\color{blue}67.1} & 66.9 \\
    \text{ViT-B/16(DeiT)} & 66.8  & 65.1 & 67.1  & {\color{blue} 68.6} & 65.8 & 66.8  & 67.3 & 64.6 & {\color{blue}68.6} & 67.1 \\ 
    % \text{ViT-B(DINO)} & x  & x & x  & x & x & x  & x & x & x & x & x\\ 
    % \text{ViT-B(DeiT)} & x  & x & x  & x & x & x  & x & x & x & x & x\\ 
    % \text{ViT-B(21k)} & x  & x & x  & x & x & x  & x & x & x & x & x\\ 
    \bottomrule
\end{tabular}}
\caption{\label{table_fewshot_orbit_ViT-B}  \textbf{Results for \orbit{} on ViT-B/16 architectures with \protoaug{} + LITE\cite{lite_meml}}.}
\vspace{-0.3cm}
\end{table*}
\section{PyTorch like Implementation of our Strong Baselines}
In~Algo.(\ref{attnpseudo}), we provide the pseudo-code for \attn{}. Note that \attn{} learns a set of scaling parameters for the attention matrix and a domain residual adapter  which is a light-weight learnable vector added to each residual connection in the ViT.  In~Algo.(\ref{attnpseudo}) the modifications to the existing code are marked in {\color{blue} blue}. It can be observed that our strong baseline \attn{} only requires minimal modifications to the existing ViT. 

In~Algo.(\ref{lnormpseudo}) -- we provide the pseudo-code for \lnorm{}. In particular, \lnorm{} requires only setting the requires$\_$grad option for LayerNorm to be set to True during fine-tuning. 
\begin{algorithm}[ht]
\SetAlgoLined
    \PyComment{... Inside init() for the ViT } \\ 
    \PyComment{nd: depth, nh: number of heads, nt: number of tokens} \\
    \PyCode{{\color{blue} scale$\_$param = nn.Parameter(torch.ones(nd,nh,nt,nt))}} \\
    \PyComment{... Inside Attention Block} \\
    \PyComment{Compute Attention} \\
    \PyCode{attn = (q @ k.transpose(-2,-1)*self.scale} \\
    \PyComment{Expand along size of support size b} \\ 
    \PyCode{{\color{blue} attn$\_$scale = scale$\_$param[i].expand(b,-1,-1,-1)}} \\ 
    \PyComment{Modification in Attention Block} \\ 
    \PyCode{{\color{blue}attn = attn$\_$scale * attn}} \\
    
    \PyComment{... Inside forward() for the ViT} \\
    \PyCode {i=0} \\
    \PyCode{for blk in blocks:} \\
    \Indp   % start indent
        \PyComment{Pass the appropriate scale param} \\
        \PyCode{x = blk(x, {\color{blue} scale$\_$param[i]})}  \\ 
        \PyCode{i += 1} \\
    \Indm % end indent, must end with this, else all the below text will be indented
    \PyComment{End} \\
\caption{\label{attnpseudo} PyTorch-style pseudocode for \attn{}}

\end{algorithm}

\begin{algorithm}[ht]
\SetAlgoLined
    \PyCode{m = model()} \\ 
    \PyComment{Set require-grad to true} \\
    \PyCode{for name, param in m.parameters():} \\
    \Indp   % start indent
        \PyComment{Check for LayerNorm param} \\
        \PyCode{if 'norm' in name:}  \\ 
        \Indp 
            \PyCode{param.requires$\_$grad=True} \\
        \Indm 
        
    \Indm % end indent, must end with this, else all the below text will be indented
    \Indp 
    \PyCode{else:} \\
            \Indp 
                \PyCode{param.requires$\_$grad=False} \\
            \Indm 
    \Indm 
    \PyComment{End} \\
\caption{\label{lnormpseudo} PyTorch-style pseudocode for \lnorm{}}
\end{algorithm}
\section{Ablation Studies}
\textbf{Only DRA.} In our experiments, we find that learning only the domain residual adapter results in inferior performance on \mdshort{}. With ViT-S/16(DINO), we find that \attn{} results in 79.4$\%$ whereas learning only the DRA module results in 76.1$\%$ on average across all the downstream domains in \mdshort{}. This result highlights that learning the set of scaling parameters in conjunction with the domain residual adapter(DRA) is crucial in obtaining strong few-shot performances. 

\textbf{\attn{} + \lnorm{}.} We also combine \attn{} and \lnorm{} to find if few-shot performance can be improved. Empirically, with ViT-S/16(DINO), we find the performance to be 79.1$\%$ -- which is similar to using \lnorm{} independently and less than $\attn{}$. This shows that combining both of our strong baselines do not provide significant advantages and should be used separately. 
%%-----------------------------------
\section{Fine-tuning Speedup Analysis}
\begin{figure}[H]
    \hskip 0.0cm
  \includegraphics[width=8.5cm, height=6.2cm]
  {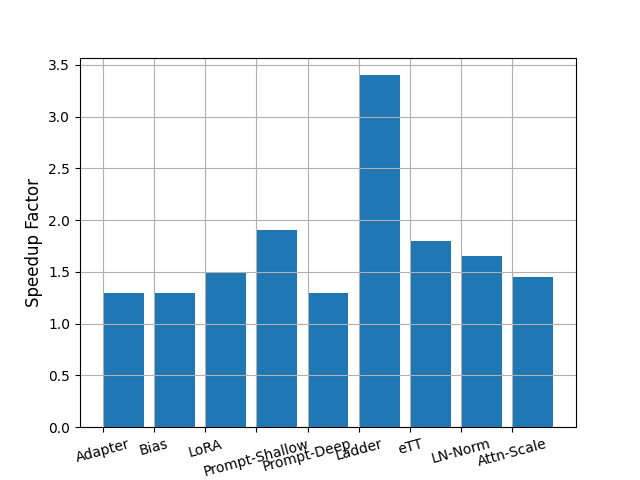}
 \caption{\label{speedup} \textbf{Ladder-tuning has a speedup factor of 3.3x, while other PEFT methods have speedup factor between 1.3-1.9x when compared to full model fine-tuning. }}
 \vspace{-0.0cm}
\end{figure}
\section{Additional Results on \orbit{}}
Due to large support set sizes in \orbit{}, we integrate LITE\cite{lite_meml} with \protoaug{}, so that fine-tuning runs can be evaluated on larger ViT architectures such as ViT-B/16. From~\Cref{table_fewshot_orbit_ViT-B}, we find that when \lnorm{} leads to the best few-shot performance, while \attn{} is competitive when compared to existing PEFT methods. Understanding the effectiveness of our strong baselines on larger ViT architectures without using the LITE framework is a direction of future work. 
\section{Note on Results with \ncc{} and \linear{}}
In general, we find that \ncc{} and \linear{} perform worse than \protoaug{} on \mdshort{} across all the pre-trained models. Although we run experiments with \ncc{} and \linear{}, we report results only with \protoaug{} due to its superior performance at the same computational cost as \ncc{}. 
\section{Further results on \attnlite{}}
In the main paper Table.(1), we find that \attnlite{} results in similar performance to \lnorm{} and slightly worse than \attn{}. In~\Cref{attn_lite_results} -- we find that for self-supervised ViT-B/16(DINO), \attnlite{} is competitive. However for supervised ViTs, it lags behind both \lnorm{} and \attn{} by a few percentage points. Improving \attnlite{} for supervised ViTs is a direction for future research.
\begin{table}[H]
\vspace{-0.2cm}
\hskip 0.2cm
\scalebox{0.8}{
\begin{tabular}{SSSS} \toprule
    {Method} & {\lnorm} & {\attn{}} & {\attnlite{}}\\ \midrule
    \text{ViT-B(DINO)} & 79.8 & 79.9 & 79.5\\ \midrule
     \text{ViT-S(DeiT)} & 77.2 & 76.3 & 75.9 \\  
    \text{ViT-B(DeiT)} & 79.0 & 78.0 & 77.1 \\
    \text{ViT-B(21k)} & 80.2 & 77.8 & 77.2 \\
  \bottomrule
\end{tabular}}
\caption{\label{attn_lite_results} \textbf{\attnlite{} is competitive for ViT-B(DINO), but lags behind \lnorm{} by a significant margin for supervised ViTs.}}
\end{table}
\section{Note on Total Number of Experiments}
We schedule experiments at the scale of: (num$\_$PEFT).(num$\_$pretrained).(num$\_$layers).(num$\_$ftalgo) = 10*5*12*3 = 1.8k experiments. 

\section{Acknowledgements}
This project was supported in part by Meta grant.
$23010098$, NSF CAREER AWARD $1942230$, $HR00111950026$ (GARD). 
This project was supported in part by Meta grant 23010098, NSF CAREER AWARD 1942230, HR001119S0026 GARD, ONR YIP award N00014 22 1 2271, Army Grant No W911NF2120076 and the NSF award CCF2212458.
\end{document}